\documentclass[conference]{IEEEtran}
\IEEEoverridecommandlockouts
\usepackage{cite}
\usepackage{amsmath,amssymb,amsfonts}
\usepackage{algorithmic}
\usepackage{graphicx}
\usepackage{textcomp}
\usepackage{xcolor}
\usepackage{subcaption}
\def\BibTeX{{\rm B\kern-.05em{\sc i\kern-.025em b}\kern-.08em
		T\kern-.1667em\lower.7ex\hbox{E}\kern-.125emX}}
		\usepackage{geometry}
\begin{document}
	
	\title{FSSC: Federated Learning of Transformer Neural Networks for Semantic Image Communication}
	
	\author{
		\IEEEauthorblockN{
			Yuna Yan\IEEEauthorrefmark{1}, 
			Xin Zhang\IEEEauthorrefmark{1}, 
			Lixin Li\IEEEauthorrefmark{1}, 
			Wensheng Lin\IEEEauthorrefmark{1}, 
            Rui Li\IEEEauthorrefmark{2},
            Wenchi Cheng\IEEEauthorrefmark{3},
			and Zhu Han\IEEEauthorrefmark{4}} 
		\IEEEauthorblockA{\IEEEauthorrefmark{1}School of Electronics and Information, Northwestern Polytechnical University, Xi’an, China, 710129}
        \IEEEauthorblockA{\IEEEauthorrefmark{2}Samsung AI Center, Cambridge, UK}
        \IEEEauthorblockA{\IEEEauthorrefmark{3}State Key Laboratory of Integrated Services Networks, Xidian University, Xi'an, China, 710071}
		\IEEEauthorblockA{\IEEEauthorrefmark{4}Department of Electrical and Computer Engineering, University of Houston, Houston, TX, 77004}
	} 
	
\maketitle

\begin{abstract}
In this paper, we address the problem of image semantic communication in a multi-user deployment scenario and propose a federated learning (FL) strategy for a Swin Transformer-based semantic communication system (FSSC). Firstly, we demonstrate that the adoption of a Swin Transformer for joint source-channel coding (JSCC) effectively extracts semantic information in the communication system. Next, the FL framework is introduced to collaboratively learn a global model by aggregating local model parameters, rather than directly sharing clients' data. This approach enhances user privacy protection and reduces the workload on the server or mobile edge.
Simulation evaluations indicate that our method outperforms the typical JSCC algorithm and traditional separate-based communication algorithms. Particularly after integrating local semantics, the global aggregation model has further increased the Peak Signal-to-Noise Ratio (PSNR) by more than 2dB, thoroughly proving the effectiveness of our algorithm. 

\end{abstract}


\section{Introduction}
Wireless communication is embracing an explosive growth of data traffic semantic nature, generated by various contemporary multi-media applications such as AR/VR, video streaming, gaming and telehealth. Traditional communication systems take the source messages (e.g. images and texts) and first encode these messages with source coding algorithms on the application layer, in order to remove redundant information. Subsequently, when the compressed information reaches the physical layer, it is then further channel-encoded by injection of redundant bits, to be resilient to the noise in the communication channel. There are three major factors that make this current approach inefficient in the 5th Generation Mobile Communication Technology(5G) and future generations of mobile networks. Firstly, although modern channel coding algorithm on its own has theoretically reached the Shannon limit on the physical layer \cite{ref1}, when jointly considering the source coding and channel coding, the Shannon's optimality theory only holds when the message blocks are asymptotically infinitely long, which is impractical in the real-world \cite{ref2,ref3,ref4,ref5,ref6}. Secondly, the separation-based approach\cite{ref7} compresses and then upscales messages, thus inherently introducing extra latency, which lead to inferior performances for major time-sensitive 5G/6G and upcoming applications, e.g. tactile internet and autonomous driving. Third, channel coding strategies treat all bits of information equally, missing the opportunity to utilize the abundant semantic information available in the messages, especially in image and video contents.

To address these issues in the traditional communication regime, semantic communication\cite{ref8,ref9} a.k.a. joint source and channel coding, has been proposed and studied, which focuses on the transmission of meaning rather than bits, helping to escape the ``Shannon trap''\cite{ref10}. 
Compared to conventional communication methods, semantic communication is centered around the relevant semantic features essential for transmitting information, thereby eliminating unnecessary redundancy. This approach leads to a significant reduction in data transmission latency and bandwidth consumption. To enhance the versatility of semantic communication systems, we propose the employment of a Swin Transformer-based semantic communication (STSC) system that primarily focuses on the reconstruction of global image signals. Its unique architecture enables the image-oriented semantic communication system to effectively capture deeper semantic features from potential semantic information and represent complex semantic relationships within the input data. In addition, the Swin Transformer is capable of reducing computational complexity and speeding up both model training and inference compared to traditional Transformer models. As a result, it is well-suited for handling large-scale semantic communication, especially in multi-user deployments.

\begin{figure*}[!t]
	\centering
	\includegraphics[width=6in]{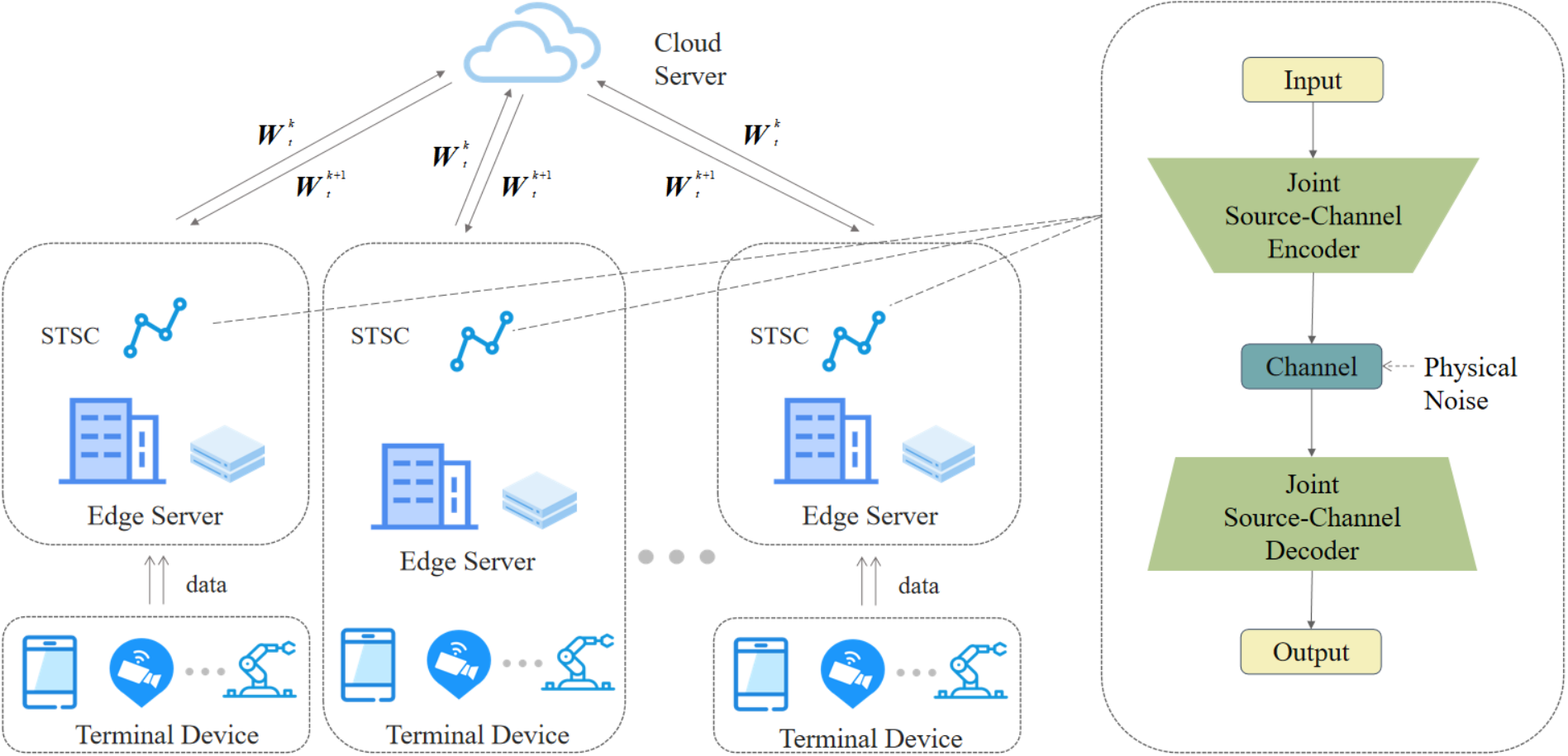}
	\caption{The overall architecture of the proposed FSSC system.}
	\label{fig_1}
\end{figure*}

However, image-oriented semantic communication targets applications that often involves users' private data, such as their faces, home interior, license plates, etc., which cannot always be uploaded to the cloud server. Moreover, most of the prior art on image-based semantic communication are point-to-point systems, which do not fit in many applications where multiple users are involved. In this paper, we propose a multi-user federated learning (FL) framework for semantic communication, showcasing the potential of efficient learning of a global semantic communication model that combines the learned knowledge from multiple users, without direct access to users' private data. We demonstrate that the proposed FL semantic system, when powered with a hierachical transformer architecture, i.e. the Swin Transformer, achieves a better image transmission quality at a faster convergence rate compared to the single-user.

In summary, the main contributions of this paper are summarized as follows:

\begin{itemize}
	\item{We propose a Swin Transformer based semantic communication architecture, which aims at minimizing the difference between the transmitted and received images, so as to effectively communicate image semantic information.}
	\item{In contrast to the centralised approach of sharing the user data directly to a cloud server in order to train a transformer neural network for semantic communication, we consider a cross-silo FL setting that employs multiple users to train locally their own model using local data, and share the local updates with a global server, thereby preserving users' data-privacy.}
	\item{Through the use of well-established dataset for computer vision, simulation results show that our proposed framework surpasses the representative JSCC methods, and further boosts peak signal-to-noise ratio (PSNR) by over $2$dB on top of aggregating local semantics.}
\end{itemize}

The rest of this paper is organized as follows. In Section \ref{sec_2}, we introduce the framework of the image semantic communication system based on FL and formulate the corresponding research problem. We then present the federated learning strategy for a Swin Transformer-based semantic communication (FSSC) in Section \ref{sec_3}. In Section \ref{sec_4}, we present numerical simulation results and showcase the semantic communication performance of FSSC as well as the convergence of the system. Finally, the paper is concluded in Section \ref{sec_5}.

\section{System Setup}
\label{sec_2}
In this paper, we consider a multi-user FL semantic communication system, as shown in Fig. \ref{fig_1}, which consists of a cloud server and multiple clients. An FSSC client can be an edge server or a terminal device. When the client is an edge server, it is able to perform computational tasks, hence the training and inference of STSC happens on the client locally. When the client is a light-weight terminal device equipped with only limited or no computational power, the client should be served by a corresponding edge server which is trusted by the client to share data with in order to perform training. The federated training process is detailed as follows:
\begin{enumerate}
	\item The cloud server initialize the STSC model for semantic communication, and distribute it to all participating clients. Each client will then perform $n$ epochs of local training using its own data. Once the local training is completed, all clients transmit the updated model parameters to the cloud server, and the server will then aggregate the updated model parameters following Federated Averaging \cite{ref13}, which completes one round of the federated learning. This process is repeated until the model converges;
	\item During deployment of a trained STSC model, a user will initiate a communication request and sends a semantic message to the client. The client will then process this message by performing an inference using the trained STSC model with the message sample as the joint source-channel encoding process, and then transmit the output of the STSC model through the uplink channel to the receiver;
	\item The transmitted message will then go through a noisy channel which typically add noisy corruption to the message. Once the receiver receives the message, it will then use the decoding part of the STSC model to recover the message. Note that the receiver in this setting could be either a cloud server, or a client.
\end{enumerate}

\begin{figure*}[!t]
	\centering
	\includegraphics[width=6.5in]{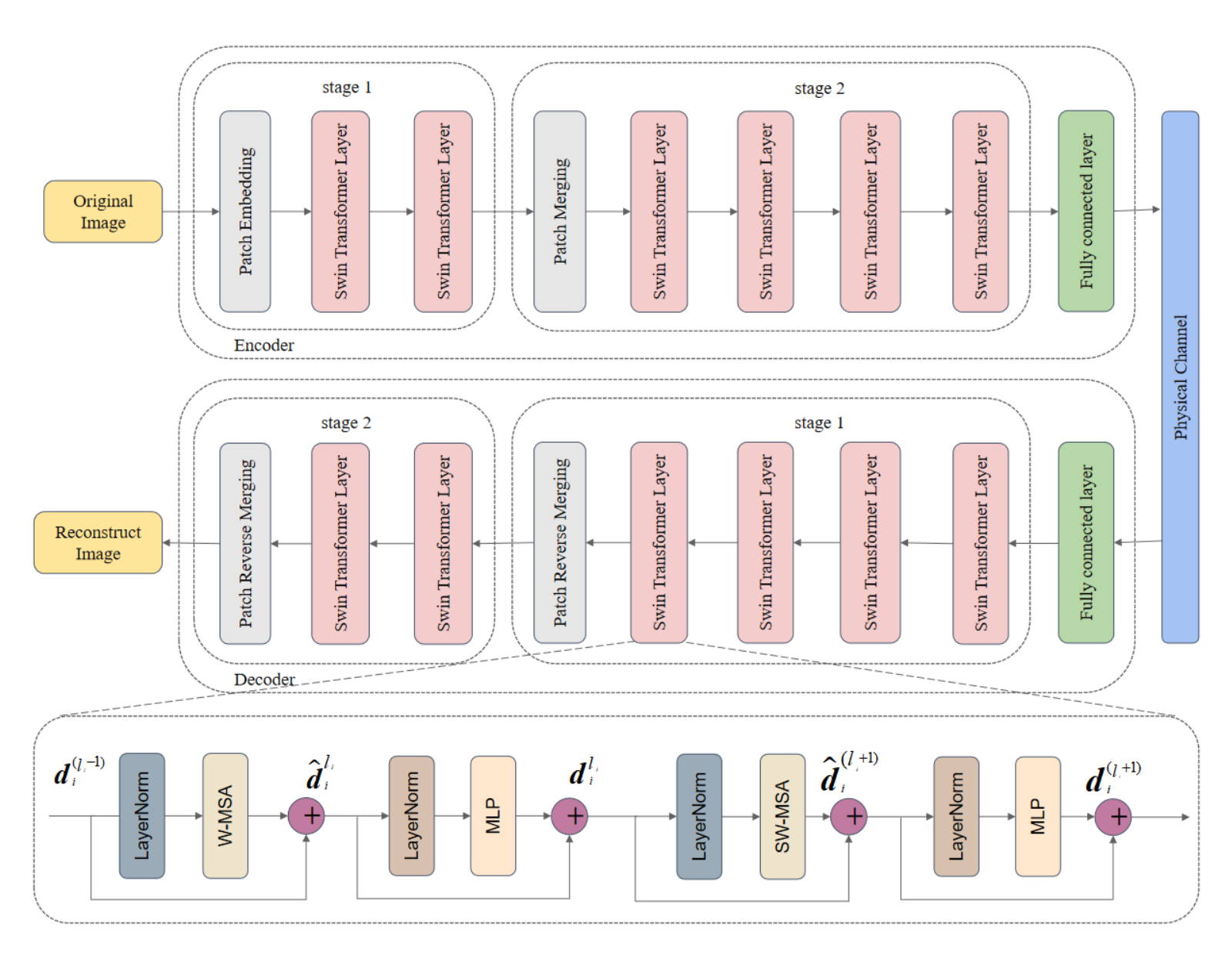}
	\caption{An illustration of the FSSC system structure with three major components: the encoder, the channel, and the decoder.}
	\label{fig_2}
\end{figure*}

Obviously, semantic information extraction and reconstruction of images are considered to be the key to the success of this system. Therefore, in this paper, the average mean square error (MSE) between the original input image and the reconstructed image is used as the loss function, which is defined as,

\begin{equation}
	\label{Eq_2}
	MSE=\frac{1}{L} \sum_{i=1}^n\left(\boldsymbol{x}_i-\boldsymbol{\hat{x}}_i\right)^2,
\end{equation}
where $ \boldsymbol{x}_i $ is the original image vector, $ \hat{x}_i $ is the reconstructed image vector, and $ L $  represents the length of the image vector. A smaller MSE indicates that the reconstructed image is closer to the original image and the quality of image reconstruction is better.

\section{Image Semantic Encoder and Decoder Network}
\label{sec_3}
In this section, we first present an edge architecture for Swin Transformer-based semantic communication (STSC), which aims to capture long-term hierarchical semantic information from images. Additionally, to enhance the accuracy of semantic information extraction and ensure user privacy, we introduce the federated learning strategy for a Swin Transformer-based semantic communication (FSSC), which enables the training of edge models using FL across multiple devices hosted on cloud servers. As a result, our proposed scheme can effectively learn semantic representations from images transmitted by different users, and ensure reliable communication on various channels.

\subsection{STSC Model}
Fig. \ref{fig_2} shows the semantic communication network structure based on Swin Transformer. In this paper, joint source channel coding is adopted for semantic communication, in which the Swin Transformer module  \cite{ref11,ref12} is adopted for encoding and decoding. Meanwhile, a non-trainable fully connected layer is used to simulate the physical channel.

At the transmitter, a set of $ N $ images $\mathbf{X}=\left\{\boldsymbol{x}_i\right\}_{i=1}^N$ is given, where $\boldsymbol{x}_i \in \mathbb{R}^{3 \times H \times W}$ denotes the $ i $-th image, and $ H $ and $ W $ denote height and width of the image, respectively. Firstly, the images are divided into non-overlapping patches through patch partion module. The size of patch is $ 4 \times 4 $  and the vector dimension is converted to $(H / 4, \mathrm{~W} / 4,48)$. 
Then, the feature block sequence is linearly embedded. And through the learnable embedding matrix $ \boldsymbol{E} $, the feature blocks can be projected into an  embedding representation of arbitrary dimension $ C $. After passing through the Swin Transformer block, the vector dimension is transformed to  $(H / 4, \mathrm{~W} / 4,C)$, 
and the whole process is called Stage 1. After the embedded representation, the feature blocks are fed into the Swin Transformer module, as shown in Fig. \ref{fig_2}. As can be seen from the figure, the first patch merging layer and Swin Transformer block are combined into Stage 2. The patch merging layer concatenates each group of adjacent patches of size $ 2 \times 2 $, so that the number of patch tokens becomes $ 1 / 4 $ of the original, i.e., $ H / 8\times  W / 8 $. Meanwhile, the dimension of patch token is expanded by four times, i.e., $ 4C $. In order to reduce the output dimension and realize the downsampling of the feature map, the patch merging layer performs a fully connected operation (implemented by a $ 1 \times 1 $  convolutional layer). After this process, the dimension of the concatenated feature patch is reduced from $ 4C $ to $ 2C $. Then, the patch goes through the swin Transformer block for feature transformation. Finally, the output dimension becomes  $(H / 8, \mathrm{~W} / 8,2C)$. The output $\tilde{\boldsymbol{x}}$ passes through a fully connected layer to facilitate transmission in the channel. Specifically, it can be expressed as,
\begin{equation}
	\label{Eq_3}
	\boldsymbol{s}=\boldsymbol{W}_1 \tilde{\boldsymbol{x}}+\boldsymbol{b}_1,
\end{equation}
where, $\boldsymbol{W}_1$ is the weight parameter matrix of the full connection, and  $ \boldsymbol{b}_1 $ is the bias of the full connection.

In order to realize the joint training of encoder and decoder, a layer of fully connected neural network is used to simulate the channel. The neural network is actually a pair of input-output mappings, where the mapping relationship is determined by the neuron weight $\boldsymbol{W}_n$ and the bias $\boldsymbol{b}_1$. Here, the weight $\boldsymbol{W}_n$ represents the channel gain, and the bias $\boldsymbol{b}_n$ is a random variable added to simulate the noise in the channel. The variance of the variable corresponds to the power of the channel noise, and its value mainly depends on the SNR and transmission power. After transmission in the physical channel, the signal $ \boldsymbol{s} $ receives interference from channel noise and becomes the signal $ \boldsymbol{y} $ when it reaches the receiver. 

At the receiver, the decoder consists of the Swin Transformer blocks. The function is to decode the signal $ \boldsymbol{y} $ to restore the signal $ \hat{\boldsymbol{x}} $.

\subsection{FL Training}
In this paper, we develop the FSSC algorithm, by applying STSC to a FL framework, which can extract semantic information accurately and carry out semantic communication while protecting user privacy.While Fig. \ref{fig_1} depics the overal architecture of FSSC, we detail below the specific training process of FSSC algorithm.

We use Fedavg\cite{ref13} as the aggregation algorithm for our federated training. Assume that $N$ clients participate in FL training. $ \boldsymbol{D}_k $  denotes the local Dataset of client $ E_k $, and $ \left|\boldsymbol{D}_k\right| $ is the size of the corresponding dataset. Thus, the training loss of the client based on the model in the local sample is
\begin{equation}
	\label{Eq_11}
	\operatorname{Loss}^k(\omega)=\frac{1}{\left|\boldsymbol{D}_k\right|} \sum_{i=1}^{\left|\boldsymbol{D}_k\right|} \operatorname{Loss}_i^k(\omega),
\end{equation}
where  $ \operatorname{Loss}^k(\cdot) $ is the training Loss of the local client $ E_k $, and $ \operatorname{Loss}_i^k(\cdot) $ is the loss of the sample $ \boldsymbol{x_i} $ corresponding to the client. In the FSSC algorithm, $ \operatorname{Loss}(\cdot) $ is represented in (\ref{Eq_2}).

Therefore, the global federated training loss can be obtained by weighted average of the loss of each client according to the dataset size. The specific formula is as follows.

\begin{equation}
	\label{Eq_12}
	\operatorname{Loss}^{\circ}(\omega)=\sum_{k=1}^N \frac{\left|\boldsymbol{D}_k\right|}{\left|\boldsymbol{D}_o\right|} \operatorname{Loss}^k(\omega),
\end{equation}
where $ \operatorname{Loss}^{\circ}(\omega) $ is the global loss after cloud aggregation, and $ |\boldsymbol{D}_o| $ is the global dataset size. The purpose of the FL training is to find the global optimal model that minimizes the sum of training losses over all data. The specific federated training process is as follows.

\textbf{Initialization.} The dataset is non-uniformly split and distributed on each client. The semantic communication network based on Swin Transformer is initialized on the cloud server, and the initialized global model parameters $ \boldsymbol{W}^o $ are broadcast to each client to build a local STSC network.

\textbf{Model training.} In the communication round $ t $, the client participating in the training will conduct local training on STSC by using the local data set. And the stochastic gradient descent method is used to update the local model parameters. For each batch $i$, the process can be expressed as,

\begin{equation}
	\label{Eq_13}
	\omega_{i+1}^k \leftarrow \omega_i^k-\eta \frac{\partial \operatorname{Loss}\left(\omega_i^k ; b\right)}{\partial \omega_i^k},
\end{equation}
where  $ \eta $ is the learning rate. Each client saves its own trained local parameters and uploads them to the cloud server.

\textbf{Parameter aggregation.} The cloud server aggregates model parameters from different clients, and updates $ \boldsymbol{W}_{t}^o $ to $ \boldsymbol{W}_{t+1}^o $ . The aggregation mechanism shown in the following equation. Through this process, the update of the global shared model can be completed.

\begin{equation}
	\label{Eq_14}
	w_{t+1} \leftarrow \sum_{k=1}^K \frac{\left|\boldsymbol{D}_k\right|}{\left|\boldsymbol{D}_o\right|} w_{t+1}^k,
\end{equation}
where $ K $ is the total number of clients participating in training. This aggregation mechanism provides a certain guarantee for the stability of the system from the perspective of model convergence. Specifically, a certain degree of client-side fluctuations can be accepted without affecting the final convergence of the model.

\textbf{Model convergence.} The parameters of the global model $ \boldsymbol{W}_{t+1}^o $  are downloaded to each client. Then, the next round of training begins. Repeat this step until a specified global training cycle is reached or the model converges.

\section{Experiment and Numerical Results}
\label{sec_4}

\subsection{Simulation Settings}
In this paper, we use the CIFAR-10 dataset\cite{ref14} to evaluate the performance of the proposed FSSC algorithm. This dataset contains 50,000 training images and 10,000 test images, each with a size of $ 32 \times 32 $. In addition, the number of clients is set to 3, and each client is assigned a different amount of non-overlapping training data. All clients use the same set of validation and test images. The training and testing environment are Windows 11+ CUDA 12.1, and the deep learning framework is Pytorch2.2.1.

\begin{figure}[!t]
	\centering
	\includegraphics[width=3.5in]{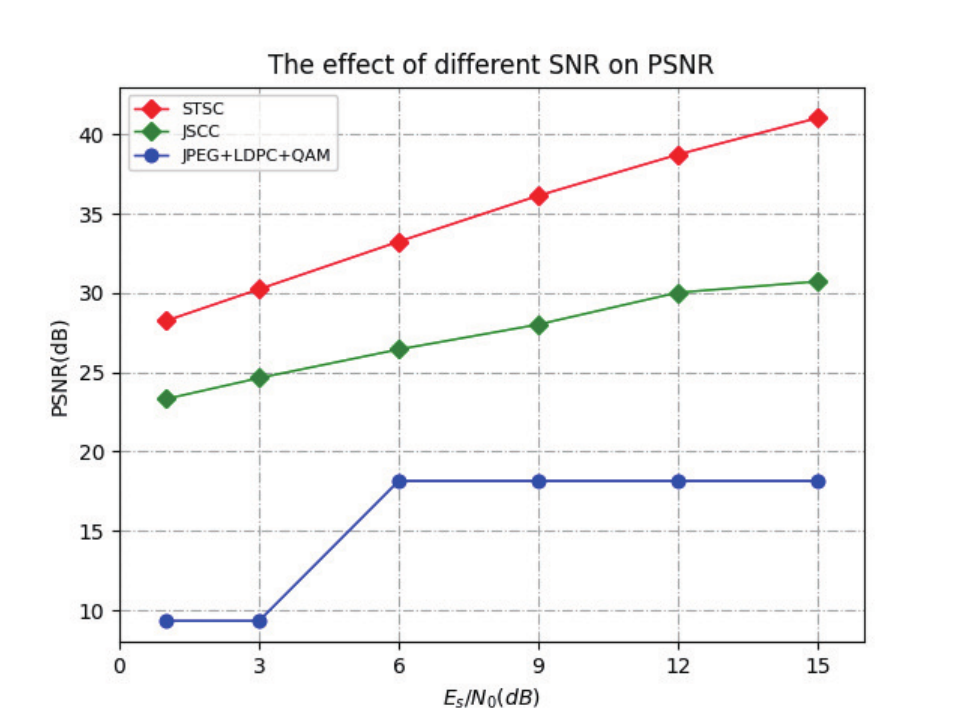}
	\caption{Comparison of the PSNR versus SNR from different methods.}
	\label{fig_3}
\end{figure}

In the experiments, the process of downloading and uploading model parameters between the client and the cloud server is regarded as an ideal situation. The focus is on the effects of global models obtained through local training and federated aggregation. The following introduces the initialization parameter settings for the semantic communication network based on Swin Transformer. Firstly, the network hyper-parameter $C$ is set to 32. The network training all uses MSE as the loss function, the number of communication round is set to 60, the batch size of the data is 64, the learning rate is 0.001, and the Adam optimizer\cite{ref15} is used. In all the tests, the compression ratio (CR) of the images is set to 0.33. 

To demonstrate the superiority of our method, we choose the typical JSCC \cite{ref2} and the traditional communication algorithm as the comparison algorithm. The traditional communication algorithm adopts JPEG \cite{ref16} as the source coding method, the channel coding adopts LDPC \cite{ref17}, and the modulation mode and the demodulation mode are Quadrature Amplitude Modulation(QAM). And The modulation order is set to 4. 

\subsection{Simulation Results}
\begin{figure}[!t]
	\centering
	\includegraphics[width=3.5in]{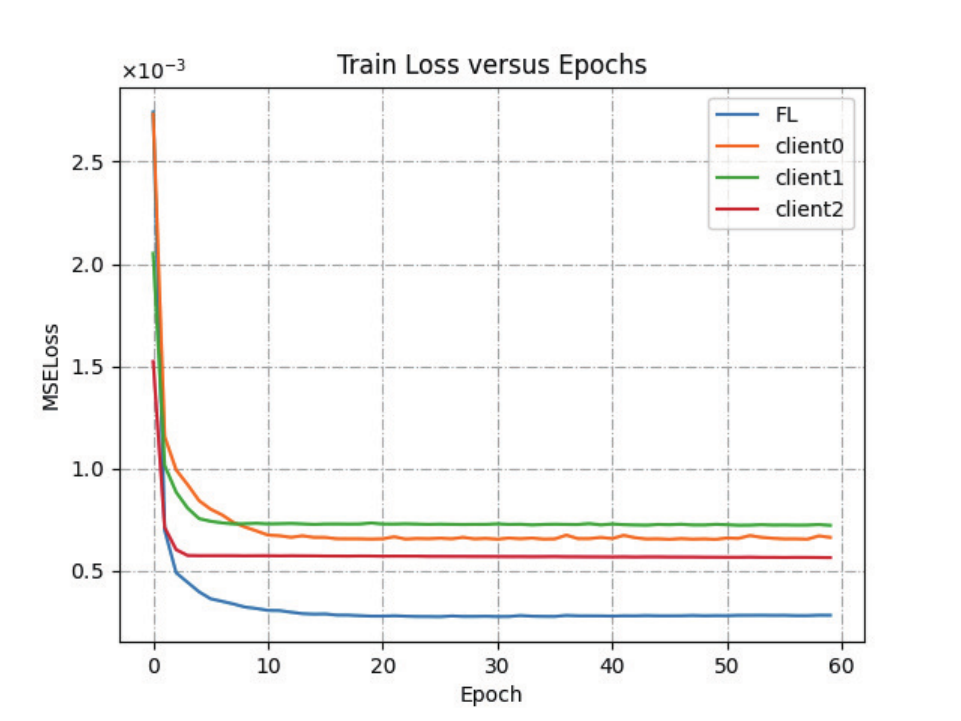}
	\caption{Convergence results of the training MSE loss versus epochs.}
	\label{fig_4}
\end{figure}

In order to evaluate the semantic representation of the FSSC model, the experiments in this section compared the performance results of the local model algorithm STSC, JSCC and the traditional communication algorithms on AWGN channels, and verified the changes of the PSNR of the three methods under different SNR conditions, as shown in Fig. \ref{fig_3}. It can be seen from Fig. \ref{fig_4} that the PSNR of STSC and JSCC gradually increase with the increase of SNR. The STSC algorithm outperformed JSCC in terms of PSNR, with an average improvement of over 5 decibels. This significant advantage demonstrates the superior efficiency of the STSC algorithm in semantic feature extraction and reconstructing image data, resulting in a more accurate representation of the original image. At the same time, it can be noted that the PSNR of JPEG+LDPC+QAM remains unchanged when the channel conditions are very bad or greatly improved. This is because when the SNR is low, the traditional communication algorithm basically cannot transmit any semantic information. However, when the SNR is high, the PSNR reaches the performance saturation of the traditional communication algorithm. Thus, the accuracy of the proposed algorithm in this paper is significantly better than that of the traditional communication algorithms, especially when the SNR is improved, the performance improvement is very obvious. This is because while LDPC codes and QAM modulation can improve the robustness of data transmission, the JPEG and BPG compression algorithm are lossy and can lead to irreversible information loss. Combining the JPEG, LDPC and QAM techniques may reduce the error resilience of the overall system, leading to degraded image quality or data integrity in the presence of transmission errors. In comparison, the STSC algorithm retains the semantic information to the greatest extent, which ensures the effective transmission of images.

In order to evaluate the convergence of the proposed STSC model under the FL framework, the convergence analysis experiment of the FSSC model is also carried out in this section. Taking $ SNR=12 $ as an example, Fig. \ref{fig_4} shows how the target loss varies as the training rounds/time increases in the case of FL training and local training, where the target loss is related to the degree of image reconstruction. As can be seen from the figure, on AWGN channels, the target loss of the communication model decreases rapidly in a short period of time, then fluctuates for a period of time, and finally remains unchanged. In other words, the model eventually reaches convergence. Moreover, the convergence efficiency of the FL training model is higher than that of the local training of the client. At the same time, the final training loss of global training model is relatively lower. This is because the local model uses a local single data set for local training, which makes it impossible to fully learn the global semantic distribution characteristics of the data. However, FL training has the advantages of data integration and improving model performance, and thus solves this problem well. 

\begin{figure}[!t]
	\centering
	\includegraphics[width=3.5in]{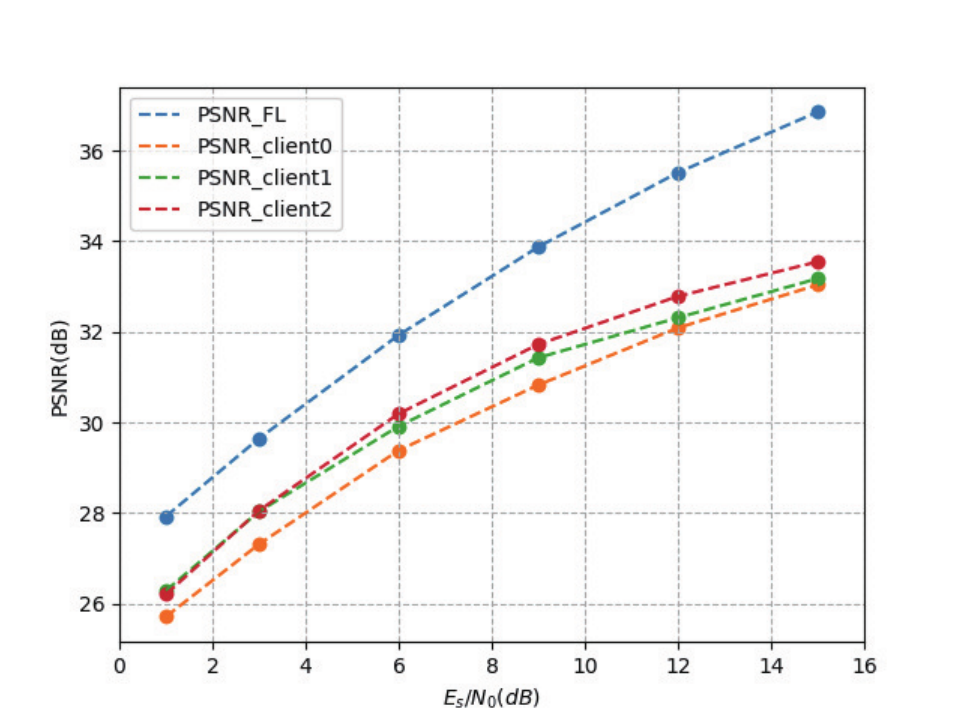}
	\caption{Comparison of the PSNR versus SNR from the global model and the local models.}
	\label{fig_5}
\end{figure}
Fig. \ref{fig_5} shows how the PSNR change with the channel SNR for the global model and the local model without federated aggregation after model convergence. From the figure, it can be observed that with the increase of SNR, the PSNR of the global model and the local model without joint aggregation also significantly improved. In contrast to local models devoid of collaborative aggregation, global models can further enhance the PSNR by 2-3dB. This is attributed to the global model's ability to learn richer noise models and finer image semantic features through effective information aggregation and synchronization. Consequently, under high signal-to-noise ratio conditions, its PSNR can continue to improve, demonstrating superior generalization and noise resistance capabilities. This is because the image reconstruction is better as the channel condition improves. Moreover,  FSSC avoids the sharing of sensitive data, thus protecting the privacy of users and motivating users to participate in semantic communication under the FL framework to a certain extent.

\section{Conclusion}
\label{sec_5}
In this paper, we proposed a semantic communication architecture for image transmission over wireless channels in multi-user cases, called FSSC. In this architecture, the local STSC semantic communication system model accurately extracts semantic information on the respective dataset and carries out the semantic communication model training. Then, joint training is performed through FL parameter aggregation to minimize the target loss function of the reconstructed image. Simulation results show that the algorithm can converge effectively. This approach is able to combine the semantic information exists in the diverse data from clients, and distributes the computational workload from edge to users, without compromising the user's privacy. Under the premise of privacy protection, the proposed algorithm can greatly reduce the amount of data required by each client while ensuring the fidelity in recovered images. Therefore, the FSSC proposed in this paper is a promising candidate scheme for multi-user image semantic communication system.

\vspace{12pt}

\end{document}